# SELF-SUPERVISED VISION-LANGUAGE PRETRAINING FOR MEDIAL VISUAL QUESTION ANSWERING

*Pengfei Li*[†]   *Gang Liu*[†*]   *Lin Tan*[†]   *Jinying Liao*[†]   *Shenjun Zhong*[‡*]

[†]College of Computer Science and Technology, Harbin Engineering University, China
[‡]Monash Biomedical Imaging, Monash University, Australia

**ABSTRACT**

Medical image visual question answering (VQA) is a task to answer clinical questions, given a radiographic image, which is a challenging problem that requires a model to integrate both vision and language information. To solve medical VQA problems with a limited number of training data, pretrain-finetune paradigm is widely used to improve the model generalization. In this paper, we propose a self-supervised method that applies **M**asked image modeling, **M**asked language modeling, **I**mage text matching and **I**mage text alignment via contrastive learning (**M2I2**) for pretraining on medical image caption dataset, and finetunes to downstream medical VQA tasks. The proposed method achieves state-of-the-art performance on all the three public medical VQA datasets. Our codes and models are available at https://github.com/pengfeiliHEU/M2I2.

## 1. INTRODUCTION AND RELATED WORK

Medical VQA task aims to generate plausible answers to clinical relevant questions with radiographic images, which requires processing information from both images and texts. Due to the absence of large-scale annotated VQA training datasets, it is challenging to train an end-to-end VQA model from scratch. Instead, early works directly adopted model weights trained using external large-scale datasets. Pretrained VGG16[1] and ResNet[2] are widely used to initialize image encoders for VQA tasks[3], while LSTM[5] and BERT[6] are used for encoding text. The dataset to pre-train those image and text encoders are non-medical image dataset, e.g. ImageNet[7], which may suffer issues of domain shifts.

To minimize the domain shifts, existing works apply pretraining to improve unimodal image encoders using only medical images. Visual encoders were commonly pre-trained with image classification tasks on external dataset and finetuned to downstream medical VQA tasks. Nguyen et al.[8] proposed a meta-learning type of pretraining task, i.e. 3-way 6-shot classification for learning visual representations using an external medical image dataset. The same group refined their solution and proposed a more data-efficient framework to select an ensemble of visual encoders trained with meta-learning setups, without external data sources[9].

Unlike the pre-trained unimodal encoders that rely on external supervised learning tasks, recent methods that design pretraining objectives with self-supervision signals to jointly model image and text features show promising results on medical VQA tasks. Masked language modeling (MLM) as a self-supervised framework was used to pre-train vision-language models[10], while incorporating visual information. Inspired by the work, contrastive language-image pretraining (CLIP)[11], Eslami et al.[12] proposed the PubMed-CLIP framework to validate the effectiveness of pretraining objectives for learning visual encoders by aligning image-text features. However, to the best of our knowledge, there are no existing works that explore both masked image and language modeling, and contrastive learning pretraining objectives in the same framework.

In this paper, we propose a self-supervised framework, that follows the "pretrain and finetune" paradigm to solve medical VQA tasks as generative processes. It includes **M**asked language modeling, **M**asked image modeling, **I**mage text matching and **I**mage text contrastive learning (**M2I2**) as pretraining objects to learn unimodal and multimodal feature representations of input image and text with image caption dataset, which is then transferred to the downstream medical VQA tasks in a data-efficient manner. Our method achieves state-of-the-art (SOTA) performance on three medical VQA datasets, VQA-RAD[13], PathVQA[14] and Slake[15], with absolute improvements of 1.3%, 13.6 and 1.1% respectively.

## 2. METHOD

In this section, we present the details of the proposed method that includes the network architectures and objectives for self-supervised pretraining and finetuning on downstream medical VQA tasks.

### 2.1. Model Architecture

In Fig. 1, it shows the network architecture in the pretraining stage. Images are encoded by a 12-layer vision transformer

(ViT)[16] and text features are extracted by a 6-layer transformer which is initialized by the first 6 layers of the pretrained BERT[6] model, while the last 6-layer BERT layers are used as the multimodal encoder for learning multimodal interaction. Similar to the Masked Autoencoder (MAE)[17], a 8-layer transformer is used as an image decoder with the weights pre-trained on ImageNet dataset[7].

In pretraining, the inputs to the model are pairs of image, V, and its corresponding captions, T. Following ViT, an image is subdivided into smaller patches of size $16 \times 16$, while the text is separated into a sequence of tokens via a BERT-based tokenizer, WordPiece. Both image and text patches are masked randomly with 15% probability, and the unmasked portion of the image and text, along with the special token <CLS> added to the beginning of the sequence, are fed into the image and text encoders correspondingly.

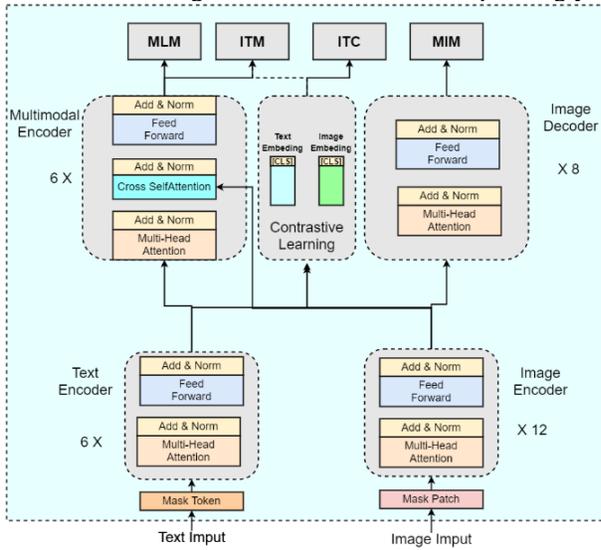

**Fig.1.** Illustration of M2I2 model.

To finetune the model to the downstream medical VQA tasks, fused multimodal features are fed into a BERT-style decoder (6 layers of transformer and a prediction head) which is for outputting the answers in a generative manner. The text decoder is initialized with the pre-trained weights. Image encoder, text encoder and the multimodal encoder are finetuned from end-to-end on medical VQA datasets, while the image decoder is not used for downstream tasks.

**2.2. Pretraining Objectives**

In this work, we include four pretraining objectives: masked image modeling and masked language modeling, image text matching and image text contrastive learning. The details are described in the following parts.

*2.2.1. Masked Image Modeling.*
To learn visual representations of images in a self-supervised manner, we apply Masked Autoencoder (MAE) which is trained to reconstruct the randomly masked image patches. A 12-layer transformer encoder with a linear layer as the prediction head is used to regress the pixel values of the masked image patches. L2 loss is used to minimize the difference between the predicted pixel values of the masked patch, and the patches in the original image, P:

$$\mathcal{L}_{mim} = \frac{1}{n}\sum_{i=1}^{n}(P_i - \widehat{P}_i)^2 \quad (1)$$

where n is the number of masked image patches.

*2.2.2 Masked Language Modeling.*
Unlike masked image modeling (MIM) which takes unimodal inputs, MLM uses both image and text information to predict the masked text token as a pretraining tasks. We randomly mask the input tokens with a 15% probability and replace the masked tokens with the special token <MASK>. The prediction is conditioned on both surrounding text information and the masked image features. The MLM loss can be defined as:

$$\mathcal{L}_{mlm} = \mathbb{E}_{(V,\widehat{T})D} H(y_{mlm}, p_{mlm}(V, \widehat{T})) \quad (2)$$

Where $\widehat{T}$ is the masked text token, $p_{mlm}(V, \widehat{T})$ is the model prediction and $p_{mlm}$ is the ground-truth of the masked text token.

*2.2.3 Image Text Matching.*
To pre-train the vision language model (VLM) using image text matching (ITM), we sample the negative text labels from the same min-batch other than the correct pair to form a binary classification task, which is similar to the other works[18]. The embedding of the <CLS> token by the multimodal encoder is treated as the joint representations of the image and text, which is then fed into a classification head. A cross-entropy loss is used for the ITM task, as below:

$$\mathcal{L}_{itm} = \mathbb{E}_{(V,T)D} H(y_{itm}, p_{itm}(V, T)) \quad (3)$$

where $H(,)$ is a cross-entropy calculation, $p_{itm}$ is the predicted class and $y_{itm}$ denotes the ground-truth label.

*2.2.4. Image Text Contrastive Learning.*
We also include the image text contrastive (ITC) learning approach proposed by ALBEF[18] as one of the pretraining objectives. The aim of the ITC task is to learn the multimodality alignment, by maximizing the similarities between the images and their corresponding text captions while pushing away from the negative ones. The image and text features are extracted using the embeddings of the <CLS> tokens by the unimodal transformer encoders respectively, and projected into 256-dimensional vector representations. The most recent image-text feature pairs are kept in the lookup table (size of 65,535). To reduce the consistency of the unimodal feature representations, we apply the momentum update trick proposed in MoCo[19]:

With all the objectives introduced above, the combined loss in our pretraining is:

$$\mathcal{L} = \mathcal{L}_{mim} + \mathcal{L}_{mlm} + \mathcal{L}_{itm} + \mathcal{L}_{itc} \quad (4)$$

### 2.3. Finetuning on Medical VQA

After pretraining using a combination of self-supervised methods on image caption dataset, we need to finetune the model to work on medical VQA tasks. As shown in Fig. 2, we initialize the image encoder, text encoder and multimodal encoder with the weights from the pretraining stage, and add the answer decoder (with the pre-trained BERT weights) to the network. The multimodal embedding output from the multi-modal encoder is fed into the answer decoder as the initial input token for the generation process. The loss used here is a conditional language-modeling loss:

$$\mathcal{L}_\theta = -\sum_{j=1}^{|y|} \log P_\theta(y_j | y < j, x, c) \quad (5)$$

where $\theta$ is the parameter of the model, $y$ represents the decoder output, $x$ is the ground-truth answers and $c$ denotes the fused context feature from the multimodal encoder.

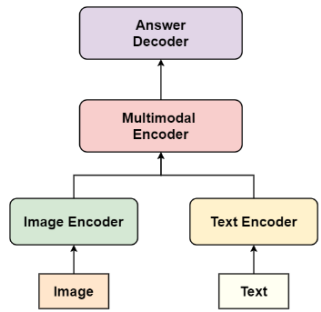

**Fig.2.** The architecture for finetuning on downstream medical VQA tasks

## 3. EXPERIMENTS AND RESULTS

### 3.1. Datasets

The ImageCLEF2022 image caption dataset[20] is used for pretraining the vision language model, which is further divided into 83,275 training and 7,645 validation samples. The model is finetune on downstream VQA tasks, using three public datasets: VQA-RAD[13], Slake[15] and PathVQA [14]. In VQA-RAD, there are totally 315 images associated with 3,064 question-answer pairs (451 pairs are selected for testing). Slake dataset is splitted into training (70%), validation (15%) and testing (15%) samples out of 14,028 QA pairs. PathVQA is the largest dataset among them, containing 32,799 QA pairs, which is divided with the proportion of 50%, 30% and 20% for training, validating and testing respectively.

There are two types of questions in the medical VQA tasks in general: closed and open ended questions. The answers to closed-ended questions have limited choice, e.g. yes/no questions. For example, "Is the mass on the left or right?". On other hand, open-ended questions do not have a limited question structure, and are more challenging tasks.

### 3.2. Experimental Details

#### 3.2.1. Pretraining
We pre-trained the model for 40 epochs on ImageCLEF2022 image caption dataset, with a batch size of 32. The images were randomly cropped into $256 \times 256$ resolution inputs and applied RandAugment[21] for data augmentation. We used AdamW optimizer with a weight decay of 0.002 and initial learning rate of $1e^{-4}$(decaying to $1e^{-5}$ by following the cosine schedule).

#### 3.2.2. Finetuning
In downstream medical VQA tasks, we finetuned the model for 40 epochs with a batch size of 8, The same optimizer, AdamW was used with a smaller learning rate, $2e^{-5}$ which was further decaying to $1e^{-6}$. Similar to ALBEF[18], the image inputs were increased from resolution of $256 \times 256$ to $384 \times 384$ in downstream tasks, and interpolated the positional encoding.

### 3.3. Results

Our approach outperformed the other methods on all the three datasets included in the evaluation, in terms of the overall performance. In Table 1, it showed that our method exceeded the current state-of-the-arts method, FITS[22], on VQA-RAD dataset, by an absolute margin of 1.3% overall (1.6% and 0.8% improvements on closed-ended and open-ended respectively). On all test set (free+para), our model was 1.7% lower than FITS in terms of closed-ended, but 1.5% and 0.3% higher in terms of closed-ended and overall.

**Table 1.** Comparisons with the state-of-the-art methods on the VQA-RAD test set. "free" means using only "freeform" answer type in test set while "free+para" means using both.

| Methods | Answer Type | Closed | Open | Overall |
|---|---|---|---|---|
| SAN-MEVF[8] | free | 74.1% | 40.7% | 60.8% |
| BAN-MEVF[8] | free | 75.1% | 43.9% | 62.6% |
| SAN-MMQ[9] | free | 73.0% | 46.3% | 62.3% |
| BAN-MMQ[9] | free | 72.4% | 52.0% | 64.3% |
| SAN-VQAMix[23] | free | 74.0% | 53.8% | 65.9% |
| BAN-VQAMix[23] | free | 79.6% | 56.6% | 70.4% |
| CMSA-MTPT[24] | free | 77.3% | 56.1% | 68.8% |
| FITS[22] | free | 80.0% | 61.0% | 72.4% |
| **M2I2(Ours)** | free | **81.6%** | **61.8%** | **73.7%** |
| MMBERT[25] | free+para | 77.9% | 63.1% | 72.0% |
| PubMedCLIP[12] | free+para | 80.0% | 60.1% | 72.1% |
| CMSA-MTPT | free+para | 80.9% | 61.5% | 73.2% |
| FITS | free+para | 82.0% | **68.2%** | 76.5% |
| **M2I2(Ours)** | free+para | **83.5%** | 66.5% | **76.8%** |

Similarly, our method outperformed the existing SOTA methods on PathVQA and Slake datasets by significant margins. As shown in Table 2, the performance of our method on the PathVQA dataset was 4.5%, 22.9% and 13.6% higher than BAN-VQAMix for closed-ended questions, open-ended and overall questions respectively. In Table 3, our method achieved 81.2% overall (which was 1.1% higher than the SOTA on that dataset, PubMedCLIP[12]), with 91.1% on closed ended and 74.7% on open-ended questions.

**Table 2.** Comparisons with the state-of-the-art methods on the PathVQA test set.

| Methods | Closed | Open | Overall |
|---|---|---|---|
| SAN-MEVF[8] | 81.0% | 6.0% | 43.6% |
| BAN-MEVF[8] | 81.4% | 8.1% | 44.8% |
| SAN-MMQ[9] | 83.7% | 9.6% | 46.8% |
| BAN-MMQ[9] | 82.1% | 11.8% | 47.1% |
| SAN-VQAMix[23] | 84.4% | 12.1% | 48.4% |
| BAN-VQAMix[23] | 83.5% | 13.4% | 48.6% |
| **M2I2(Ours)** | **88.0%** | **36.3%** | **62.2%** |

**Table 3.** Comparisons with the state-of-the-art methods on the Slake test set.

| Methods | Closed | Open | Overall |
|---|---|---|---|
| VGG+SAN[15] | 76.1% | 70.3% | 72.7% |
| VGGseg+SAN[15] | 75.4% | 72.2% | 75.4% |
| PubMedCLIP[12] | 82.5% | **78.4**% | 80.1% |
| **M2I2(Ours)** | **91.1**% | 74.7% | **81.2**% |

### 3.4. Ablation Study

**Table 4.** Ablation study on the VQA-RAD, PathVQA and Slake test set.

| Datasets | Methods | Closed | Open | Overall |
|---|---|---|---|---|
| VQA-RAD | M2I2(w/o Pretraining) | 70.8% | 36.6% | 57.1% |
| | M2I2(w/o MIM) | 74.1% | 52.0% | 65.3% |
| | M2I2(w/o ITC) | 73.5% | 45.5% | 62.3% |
| | **M2I2** | **81.6%** | **61.8%** | **73.7%** |
| PathVQA | M2I2(w/o Pretraining) | 87.4% | 26.8% | 57.2% |
| | M2I2(w/o MIM) | 87.9% | 34.8% | 61.4% |
| | M2I2(w/o ITC) | 88.0% | 36.1% | 62.1% |
| | **M2I2** | **88.0%** | **36.3%** | **62.2%** |
| Slake | M2I2(w/o Pretraining) | 83.2% | 67.3% | 73.5% |
| | M2I2(w/o MIM) | 87.0% | 72.6% | 78.2% |
| | M2I2(w/o ITC) | 86.5% | 72.7% | 78.1% |
| | **M2I2** | **91.1**% | **74.7%** | **81.2**% |

To further verify the effectiveness of pretraining objectives in our method, we conducted ablation study on all the three medical VQA datasets. As shown in Table 4, applying self-supervised pretraining was critical factor to achieve good performance on downstream medical VQA tasks, particularly for small datasets like VQA-RAD where the model trained with all 4 pretraining objectives performed significantly better than the model without pretraining by more than 16%. In the same testing dataset, a more than 25% improvement was observed for open ended questions if the model was pre-trained on image caption dataset. For relatively larger datasets, notable performance improvements were made with pretraining tasks, i.e. roughly 5% for both PathVQA and Slake dataset.

We also investigated into the contributions of masked image modeling and image-text contrastive learning techniques as pretraining tasks. In table 4, it showed that model performance without MIM or ITC dropped dramatically on small dataset, i.e. VQA-RAD, by 8.4% and 9.4% respectively. On the other hand, performance of direct training models on medical VQA datasets without pretraining declined by much smaller margins.

### 3.5. Visualization

We visualized the cross-attention maps of the models with Grad-CAM[26] and investigated the regions highlighted in the image which contributed the most to generating the answers. As shown in Fig. 3, we presented 4 attention maps overlaid on the original images, from image-question pairs of both open and closed ended types. In Fig. 3a, to answer the question, "Are regions of the brain infarcted?", the model attended on the areas of necrosis accurately. Similarly, as shown in Fig. 3c and Fig. 3d, the model focused on the correct regions for answering open ended questions. Interestingly, in the example reported in Fig. 3c where the question was "Which hemisphere are the lesions located in?", the model paid attention to the lesions on both hemispheres and output the answer, "bilateral". Besides, to answer counting-related questions, like the example in Fig. 3b where the question was "Are there >12 ribs", the model was capable of attending to the regions of ribs and recognizing the semantics of the "greater-than" sign in the question.

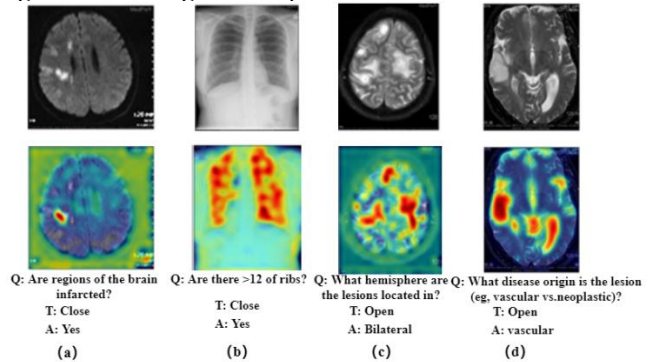

**Fig.3.** Visualizations of the image attention maps on medical VQA tasks.

### 4. CONCLUSION

In this paper, we proposed a new framework, M2I2 that learns vision-language representations by pretraining with masked image and language modeling, contrastive learning and image-text matching as training objectives. The method not only outperformed the SOTA methods by a significant margin, when finetuning to downstream medical VQA tasks, but also showed the potential of good model explainability.

### 5. ACKNOWLEDGMENTS

This work is supported by Natural Science Foundation of Heilongjiang Province under grant number LH2021F015, and National Foreign Cultural and Educational Expert Project under grant number G2021180008L.